\documentclass[10pt,twocolumn,letterpaper]{article}

\usepackage{iccv}
\usepackage{times}
\usepackage{epsfig}
\usepackage{graphicx}
\usepackage{amsmath}
\usepackage{amssymb}

\usepackage[breaklinks=true,bookmarks=false]{hyperref}

\iccvfinalcopy 

\def\iccvPaperID{****} 

\ificcvfinal\fi

\begin{document}

\title{Multi-scale Dynamic Feature Encoding Network for Image Demoir\'{e}ing}

\author{Xi Cheng, Zhenyong Fu and Jian Yang\\
PCA Lab, Key Lab of Intelligent Perception and Systems for High-Dimensional Information of \\Ministry of Education, and Jiangsu Key Lab of Image and Video Understanding for Social Security,\\
  School of Computer Science and Engineering\\
Nanjing University of Science and Technology, Nanjing, China\\
{\tt\small \{chengx,z.fu,csjyang\}@njust.edu.cn}
}

\maketitle
\ificcvfinal\thispagestyle{empty}\fi

\begin{abstract}

The prevalence of digital sensors, such as digital cameras and mobile
phones, simplifies the acquisition of photos. Digital sensors, however,
suffer from producing Moir\'{e} when photographing objects having complex
textures, which deteriorates the quality of photos. Moir\'{e} spreads
across various frequency bands of images and is a dynamic texture with
varying colors and shapes, which pose two main challenges in
demoir\'{e}ing---an important task in image restoration. In this
paper, towards addressing the first challenge, we design a multi-scale
network to process images at different spatial resolutions, obtaining
features in different frequency bands, and thus our method can jointly
remove moir\'{e} in different frequency bands. Towards solving the
second challenge, we propose a dynamic feature encoding module (DFE),
embedded in each scale, for dynamic texture. Moir\'{e} pattern can be
eliminated more effectively via DFE.
Our proposed method, termed Multi-scale convolutional network with
Dynamic feature encoding for image DeMoir\'{e}ing (MDDM), can outperform
the state of the arts in fidelity as well as perceptual on
benchmarks.
\end{abstract}

\section{Introduction}
\label{section:introduction}

\begin{figure}[t]
  \begin{center}
    \includegraphics[width=0.8\linewidth]{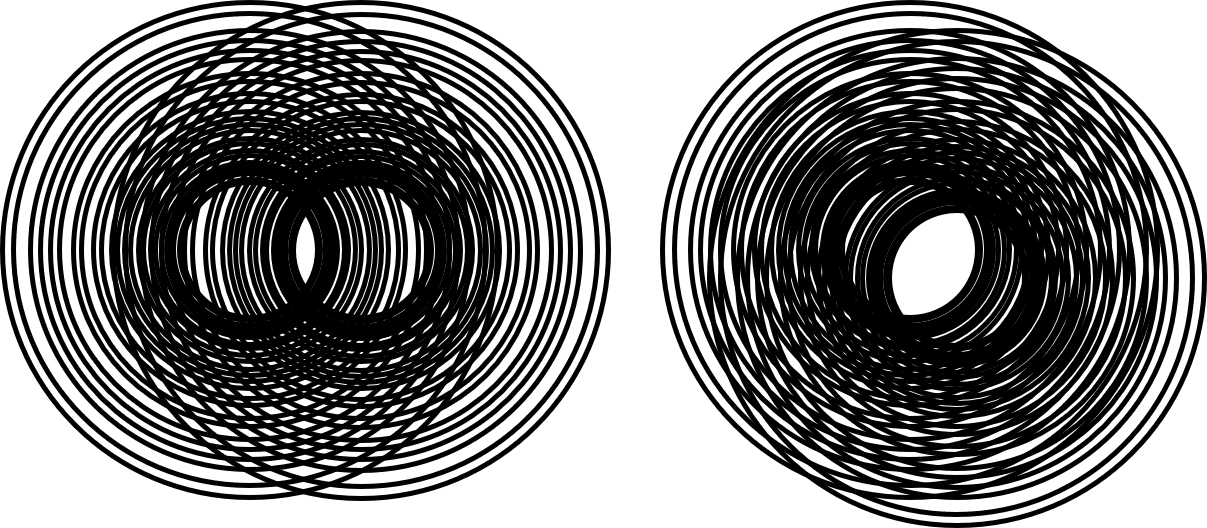}
  \end{center}
  \caption{Visualization for the formation of moir\'{e} pattern. Different misalignment will cause different moir\'{e} patterns}
  \label{fig:moire_formation}
\end{figure}

Photographing with a mobile phone or digital camera often generates
moir\'{e}, which seriously affects the visual quality of captured
images. Moir\'{e} usually appears in the form of colored stripes,
changing significantly due to the angle and distance of the
shot. Using a digital camera or mobile phone to shoot a computer or TV
screen exemplifies the moir\'{e} phenomenon where moir\'{e} appears
more obviously. Since the screen is composed of LED lattices and the
Color Filter Array (CFA) of camera's sensor also has a certain texture
structure, the imperfect alignment~\cite{moirepattern} between them
mainly causes the generation of moir\'{e} pattern. Figure~\ref{fig:moire_formation} gives a visualization
for the formation of the moir\'{e} pattern. The left and right rings in Figure~\ref{fig:moire_formation} generate different moir\'{e} patterns
result from different misalignment and show the dynamic and variable characteristics of moir\'{e} patterns.

\begin{figure}[t]
  \begin{center}
    \includegraphics[width=0.8\linewidth]{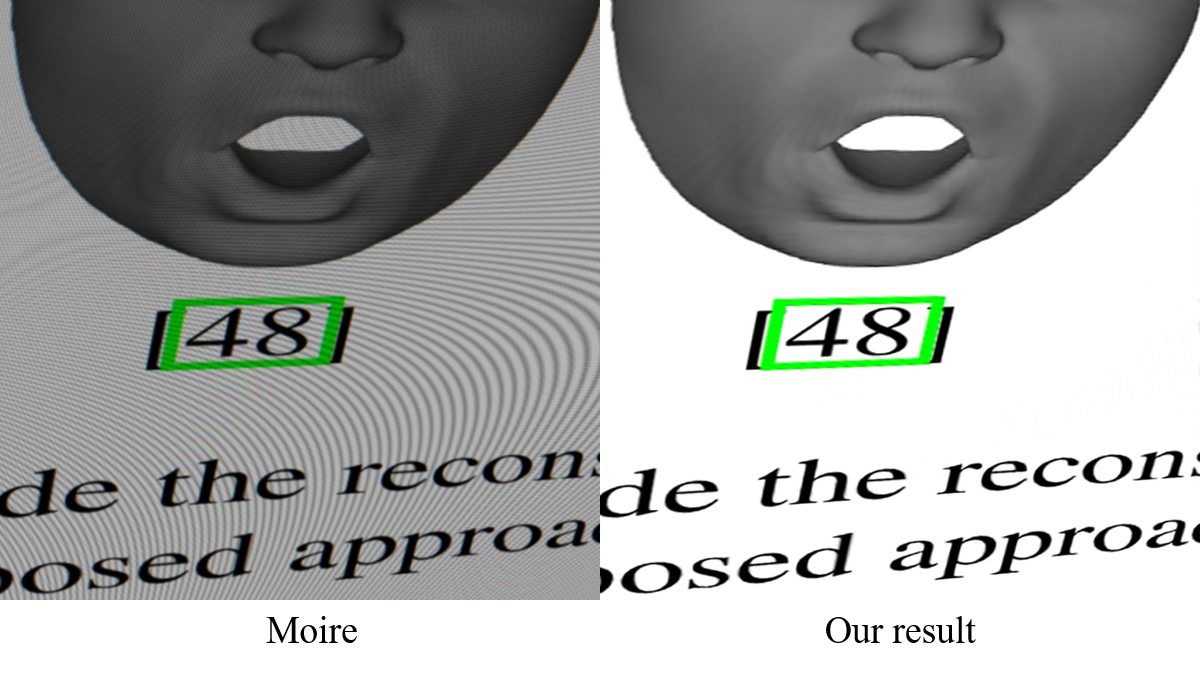}
  \end{center}
  \caption{Visual result for image 000041 from demoir\'{e}ing dataset
    (DD). Our proposed MDDM can effectively remove the moir\'{e} pattern.}
  \label{fig:visual_result}
\end{figure}

Demoir\'{e}ing differs from traditional image restoration tasks such
as denoising and super-resolution. The latter is usually static, with
relatively consistent degradation on the image, while moir\'{e} is often
dynamic and will vary with sensor resolution, distance, and
direction. Moreover, the frequency distribution of moir\'{e} is broad,
covering both the low-frequency part and the high-frequency part,
while the tasks such as image denoising and super-resolution only need
to process the high-frequency part of images. Therefore,
demoir\'{e}ing is more difficult than other image restoration tasks.

To remove the moir\'{e} pattern in images, we construct an image
feature pyramid, encode image features at different spatial
resolutions, and obtain image representations of different frequency
bands. Specifically, we propose a multi-scale residual network with
multiple resolution branches. These branches learn the nonlinear
mapping on the original resolution and the $2\times$, $4\times$,
$8\times$, $16\times$ and $32\times$ down-sampled resolutions. Then,
the features are up-sampled to the original resolution with sub-pixel
convolution~\cite{shi2016real} at the end of each branch. The network
automatically learns the weight of each branch and eventually sums
the results from different resolutions to get the final output, i.e.,
the demoir\'{e}ing image.

The moir\'{e} pattern is a dynamic texture. To encode the moir\'{e}
pattern \emph{dynamically}, we propose a dynamic feature encoding
(DFE) module. Especially, at each down sampling branch, we introduce
an extra lightweight branch. The number of convolutional layers in
this extra branch equals to the number of residual blocks in the
backbone branch. Each dynamic feature coding branch encodes the
characteristic of the global residual at different scales. In the
end, we impose the coding of the dynamic moir\'{e} pattern back to the
main branch through adaptive instance normalization
(AdaIN)~\cite{huang2017arbitrary,wang2017zm} to guide the
demoir\'{e}ing process. Figure~\ref{fig:visual_result} gives a visualization
of our demoir\'{e}ing method which effectively removes the moir\'{e} patterns in the image.

We summarize our main contribution in the following points:
\begin{itemize}
    \item We propose a novel method to remove moir\'{e} pattern through a
      progressive multi-scale residual network which enables the model
      to learn the representations on multiple frequency bands.
    \item We further propose a dynamic feature encoding (DFE) module
      that can encode the variations of moir\'{e} patterns in images
      so that the model can better cope with the variability of
      moir\'{e} pattern.
    \item The proposed method outperforms the state of the arts and
      achieves the 2nd place in the AIM2019 Demoir\'{e}ing Challenge~\cite{AIM19demoireMethods} - Track 1: Fidelity and the 3rd
	place in Track 2: Perceptual.
\end{itemize}

We organize the rest of this paper as follows: Section~\ref{section:related_works} discusses the
researches related to this article, and Section~\ref{section:proposed_method} elaborates the
structure of the proposed demoir\'{e}ing network. Section~\ref{section:experiment} shows our
experimental results. In Section~\ref{section:ablation_study}, we study the internal parameters
and structure of the network. Section~\ref{section:conclusion} concludes our paper.

\section{Related Work}
\label{section:related_works}
Image restoration is an important task in low-level vision. In image
restoration, a large number of recent researches mainly focus on tasks
such as image
super-resolution~\cite{dong2015image,cheng2019triple,dai2019second,shi2016real},
image denoising~\cite{Zhang2016Beyond}, and image
deblurring~\cite{kupyn2018deblurgan}. Demoir\'{e}ing is also an
important image restoration task and is even more difficult than
other image restoration tasks. However, in recent years,
demoir\'{e}ing has received rare attention; relatively few related
studies focus on this topic.

Traditional demoir\'{e}ing approaches are mainly based on filtering or
image decomposition methods. Wei et al.~\cite{wei2012median} proposed a
median-Gaussian filtering method for eliminating moir\'{e} in X-ray
microscopy images. Liu et al.~\cite{liu2015moire} applied a low-rank
and sparse matrix decomposition-based method to remove moir\'{e} on
texture images. Yang et al.~\cite{yang2017demoireing} proposed a layer
decomposition method based on polyphase components (LDPC) to better
remove the moir\'{e} generated when shooting the screen.

Deep neural networks have promoted many computer vision tasks greatly
compared with traditional methods. Deep convolutional neural networks
have been used in image restoration tasks in recent years. Sun et
al. proposed a demoir\'{e}ing network based on multi-scale
convolutional neural networks~\cite{sun2018moire} and a large
moir\'{e} image dataset based on
ImageNet~\cite{deng2009imagenet}. This network provided an impressive
result in removing moir\'{e} pattern. Later, Gao et al.~\cite{gao2019moire} improved the multi-scale
framework with feature enhancing branch~\cite{Zhang2018ExFuse} which enhanced semantic information with a fusion of low and high level features.
However, their proposed method do not have strong network architectures, the feature
expression ability is weak, and the characteristics of the input image
cannot be fully extracted. Moreover, they used transposed convolution~\cite{zeiler2010deconvolutional} for up sampling which
 cannot usually cannot make good use of the information and causes
artifacts in the generated images. Liu et al.~\cite{liu2018demoir} proposed a method for removing
moir\'{e} pattern based on a deep convolutional network for screenshot
images. Their proposed method could effectively remove most of the moir\'{e}
noise. However, they only used a low-resolution scale input and could only
characterize the moir\'{e} in a single frequency band. The frequency of
the moir\'{e} is widely distributed, and thus the moir\'{e} pattern cannot be well
removed in their method.

\begin{figure}[t]
  \begin{center}
    \includegraphics[width=0.8\linewidth]{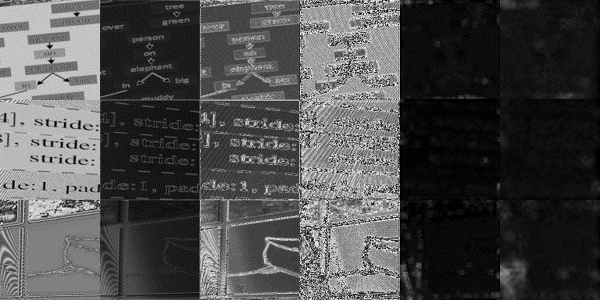}
  \end{center}
  \caption{Visualization of multi-scale results}
  \label{fig:multi_scale_visualize}
\end{figure}

\begin{figure*}
  \begin{center}
    \includegraphics[width=0.8\linewidth]{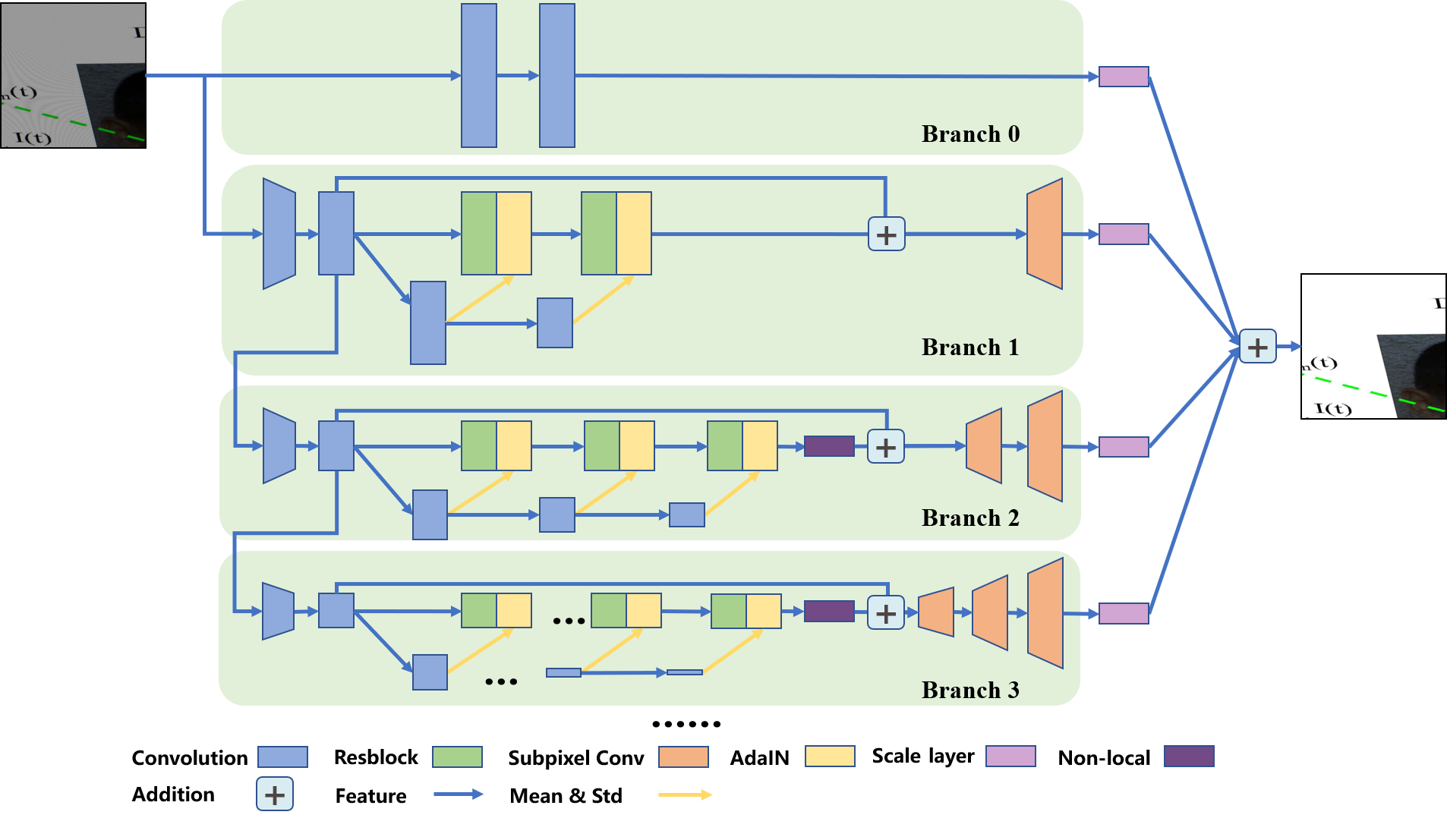}
  \end{center}
  \caption{Overall structure of our proposed demoir\'{e}ing model
    (MDDM). The input moir\'{e} image is decomposed to different
    frequency bands through different resolution branches. Dynamic
    feature encoding is used for coping with the variability of
    moir\'{e} pattern. Non-local block helps the model learn the
    region-level non-local information.}
  \label{fig:structure}
\end{figure*}

To overcome the limitations in existing demoir\'{e}ing methods, we
propose two vital schemes: a multi-scale or multi-resolution network
structure and a dynamic feature encoding module. The multi-resolution
network, aiming at processing the moir\'{e} image at multiple
resolutions, and the dynamic feature coding module, aiming at encoding
the variability of dynamic moir\'{e} patterns, benefit from each other
for removing moir\'{e}. We will detail our proposed method in the next
section.

\section{Proposed Method}
\label{section:proposed_method}
Our proposed demoir\'{e}ing network is a multi-branch structure that can
be considered as a single encoder and multi-decoder structure. We use an
encoder to downsample the input moir\'{e} image to $\frac{1}{2}$,
$\frac{1}{4}$, $\frac{1}{8}$, $\frac{1}{16}$ and $\frac{1}{32}$ of the
original resolution, encoding image features in different frequency
bands. Two key parts in our proposed method, multi-scale structure and
dynamic feature encoding will be elaborated in Section~\ref{subsection_multiscale_network} and Section~\ref{subsection:dfe}, respectively.

\subsection{Multiscale residual network}
\label{subsection_multiscale_network}
The existence of moir\'{e} often lies in the imperfect alignment
between the repetitive structure or texture of objects being
photographed, e.g. the LED array of the screen, and the Bayer array on
the camera sensor~\cite{moirepattern}. The moir\'{e} pattern is
distributed throughout the low- and high-frequency portions of an
image. Continuously downsampling the image, image representations from
high frequency to low frequency will be
separated. Figure~\ref{fig:multi_scale_visualize} visualizes the
representations of different resolutions decomposed from the original
image. If we only demoir\'{e} at the original resolution, more
high-frequent details of images will be retained, but the
low-frequency moir\'{e} pattern will be difficult to be removed
cleanly. On the contrary, although demoir\'{e}ing after downsampling
can make the image cleaner, the demoir\'{e}d image will lose more
high-frequency details. Therefore, the demoir\'{e}ing network must
learn the representations of different frequency bands at different
resolutions. After combining the demoir\'{e}d results at various
resolutions, the reconstructed image can cleanly remove moir\'{e} and
in the meanwhile maintain more image details.

To handle the demoir\'{e}ing task, we construct a six-branch fully
convolutional network containing the original resolution and $2\times$, $4\times$,
$8\times$, $16\times$ and $32\times$ down-sampled resolutions. According to the spatial
size of the input feature, the higher resolution branch will have more
computational complexity than the lower resolution branch in the same
network structure. The higher resolution branch itself retains more
high-frequency information, and the lower resolution branch needs a
deeper network structure to recover more useful high-frequency
information. Thus, we build a progressive network structure as shown
in Figure~\ref{fig:structure}. High-resolution branches have fewer
convolutional layers, and low-resolution branches have more
convolutional layers and more complex structures.

\begin{figure}[t]
  \begin{center}
    \includegraphics[width=0.8\linewidth]{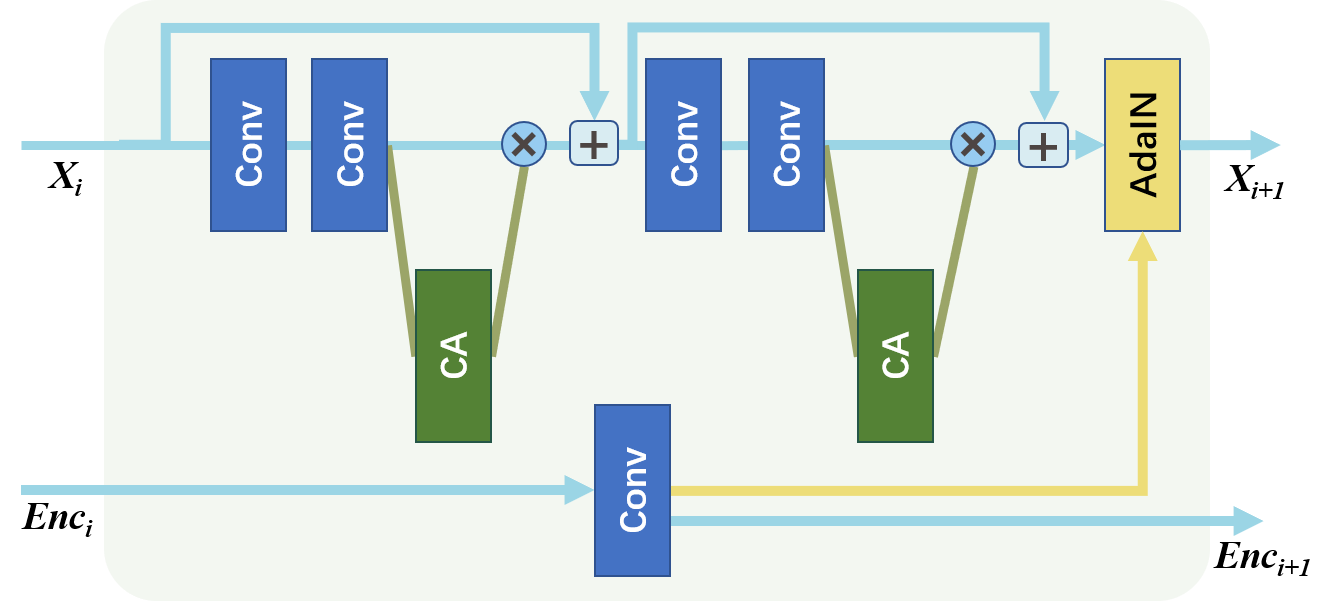}
  \end{center}
  \caption{The structure of our proposed Channel attention Dyanmic
    feature encoding Residual block (CDR).}
  \label{fig:CDR}
\end{figure}

In our model, Branch-0 remains the original resolution and only uses a
3-layer convolution structure. The calculation process of Branch-0 is:
\begin{equation}\label{eq.1}
  F_0 = \sigma{^p} W_3(\sigma{^p} W_2(\sigma{^p} W_1(I))),
\end{equation}where $F_0$ is the output feature generated from
Branch-0, $I$ denotes the input image, $\sigma{^p}$ means
PReLU~\cite{he2015delving} activation function and $W$ means the
weight of convolution layers in the branch. The resolution of Branch-1
is half of the original, using two residual blocks and increasing
channel attention. The calculation process of Branch-1 can be
expressed as follows:
\begin{equation}
\begin{aligned}\label{eq.2}
  F_1=& Up(CDR_3(CDR_2(CDR_1(Down(F_0))))\\
  & +Down(F_0)),
\end{aligned}
\end{equation}where $F_i$ and $F_{i+1}$ denote the output feature from
the adjacent branch with higher resolution and the output feature of
the current branch, respectively. $Down(.)$ is the downsample block
and $Up(.)$ is the upsample block~\cite{shi2016real}. $CDR$ is the
abbreviation of Channel attention Dynamic feature encoding Residual
block, of which the structure is shown in Figure~\ref{fig:CDR}.
The operation of channel attention can be divided into the
squeeze and excitation parts, as defined in Eq.~\ref{eq.3} and
Eq.~\ref{eq.4}, respectively.
\begin{equation}\label{eq.3}
  S(x_c)=\frac{1}{HW}\sum_{i}^{H}\sum_{j}^{W}x_c(i,j),
\end{equation}where $S(.)$ represents the squeeze process with global
average pooling~\cite{lin2013network}, $H$ and $W$ denote the height
and width of the feature map, and $x_c$ means channel $c$ of the input
feature map $x$.
\begin{equation}\label{eq.4}
  A_c(x)=\sigma{^s}(W_u\sigma{^p}(W_dS(x)))*x,
\end{equation}where $A_c(.)$ denotes the channel attention function,
$\sigma{^p}$ means the PReLU~\cite{he2015delving} function, $W_u$ and
$W_d$ denote two $1\times 1$ convolution layers containing 1/16 of the
original channels to form a bottleneck, $\sigma{^s}$ means the sigmoid
function which maps the features between 0 to 1, and we use these
features as a weight to refine the original information in the
residual.

The structures of Branch-2 to Branch-5 are similar to Branch-1, except
that the number of residual blocks is increased, corresponding to
spatial resolutions of 1/4, 1/8, 1/16, and 1/32, respectively:
\begin{equation}\label{eq.5}
  F_{i+1} = Up(NL(Down(CDR_k ... (CDR_1(F_i)+F_i))),
\end{equation}where $F_i$ and $F_{i+1}$ denote the feature from the
adjacent branch with higher resolution and the feature of the current
branch, $CDR_k$ means the $k$-th block in the branch, $NL(.)$ means
the region-level non-local operation~\cite{dai2019second} at the
end of the branch, which helps the model learn self-similarity in the
current resolution.

\begin{figure*}[h]
  \begin{center}
    \includegraphics[width=0.8\linewidth]{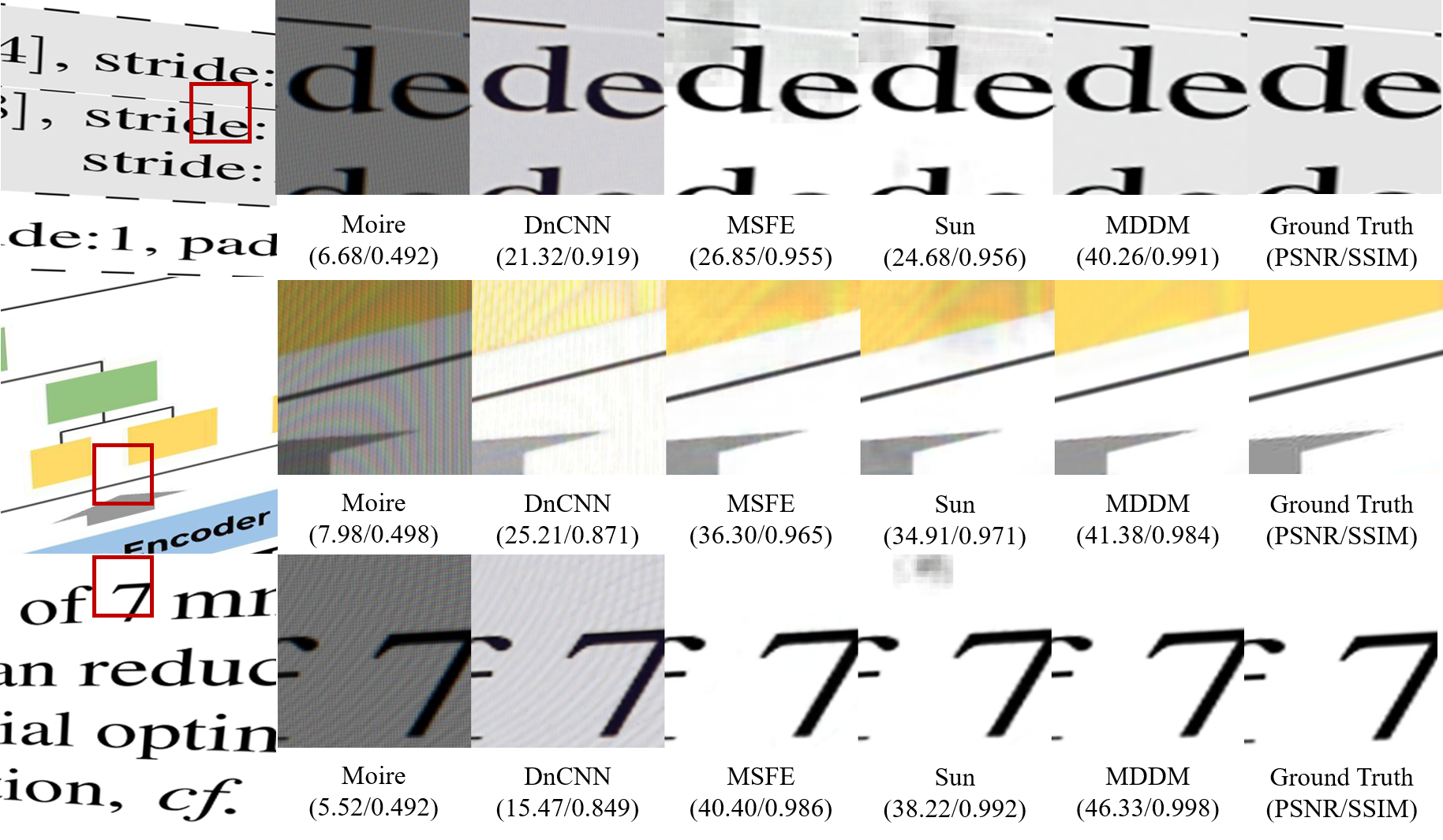}
  \end{center}
  \caption{We compare our proposed MDDM with DnCNN, MSFE and Sun.}
  \label{fig:compare_visualize}
\end{figure*}

\subsection{Dynamic feature encoding}
\label{subsection:dfe}
Demoir\'{e}ing differs from traditional image restoration tasks, such
as super-resolution and denoising. Moir\'{e} is a dynamic pattern in
that the interference texture on an image is local and varies with
scaling and angle. In contrast, the restoration pattern of
super-resolution or denoising is static, because if a specific
degeneration model, e.g. bicubic downsampling and Gaussian noise, is
used, the effect on the entire image is often consistent.

In this paper, we propose a dynamic feature encoding (DFE) module to
enhance the ability of our demoir\'{e}ing model to cope with dynamic
textures.
The global residual
learning~\cite{kim2016accurate} is used internally in the branch of
each resolution in our proposed network.
We use the residual block to model the difference between the clean and
moir\'{e} images at each feature level and frequency band, that is, the
moir\'{e} pattern on each branch.
The inconsistent nature of moir\'{e} affects the learning of
demoir\'{e}ing network, and is more intense than other image
restoration tasks.

Inspired by the recent work in arbitrary image style
transfer~\cite{huang2017arbitrary,wang2017zm}, we design a bypass branch for each
main scale branch to encode image features at different spatial
resolutions.
We embed the extra bypass branch into each backbone resolution branch
using adaptive instance normalization (AdaIN)~\cite{huang2017arbitrary}. Specifically,
in AdaIN, we first calculate the mean value and the variance of
the feature map as follows:
\begin{equation}\label{eq.6}
  \hat{\mu}_i = \frac{1}{HW}\sum_{j=1}^{H}\sum_{k=1}^{W}x_{ijk}^{enc},
\end{equation}and
\begin{equation}\label{eq.7}
  \hat{\sigma}^2_i = \frac{1}{HW}\sum_{j=1}^{H}\sum_{k=1}^{W}(x_{ijk}^{enc}-\mu_i)^2,
\end{equation}where $H$ and $W$ denote the height and width of the
feature map, $\hat{\mu}_i$ and $\hat{\sigma}^2_i$ are the mean
value and the variance of feature $x^{enc}$ from the $i$-th encoding
layer in the dynamic feature encoding branch.
After we calculate the statistical values of the moir\'{e} pattern,
we use these values to dynamically adjust the parameters of the
backbone resolution branch via adaptive instance normalization as
follows:
\begin{equation}\label{eq.8}
  x_{i+1} = \frac{x_{i}-\mu_i}{\sqrt{\sigma^2_i}+\epsilon}\sqrt{\hat{\sigma}^2_i}+\hat{\mu}_i,
\end{equation}
where
$\mu_i$ and $\sigma^2_i$ denote the statistical information from the
backbone branch, and $x_i$ denotes the feature map from the $i$-th
residual block in the backbone branch. Section~\ref{subsection:ablation_dfe} gives an
experimental analysis of dynamic feature coding.

\subsection{Reconstruction network and loss function}
\label{subsection_reconnet}
Different branches learn feature representations at different
resolutions. The distributions of moir\'{e} pattern and the image
details on different resolution branches are all different. Thus, we
design a branch scaling module that can automatically learn the
importance of each branch with backpropagation, giving each branch an
importance weight. The reconstruction process of the final
demoir\'{e}ing image is represented in Eq.~\ref{eq.9} as follows:
\begin{equation}\label{eq.9}
  \hat{I} = S_0(F_0)+...+S_6(F_5),
\end{equation}where $\hat{I}$ is the reconstructed clean
(demoir\'{e}d) image, $S(.)$ is the scaling function, $F_i$ is the
output feature of each branch.
Since directly optimizing the mean squared error (MSE) is easier to
make the image over smooth and thus blur, we instead minimize the
Charbonnier loss~\cite{lai2017deep}. Specifically, this loss function
is defined in Eq.~\ref{eq.10}:
\begin{equation}\label{eq.10}
  Loss = \frac{1}{N}\sum_{i=1}^{N}\sqrt{(\hat{I}-I)^2-\epsilon^2},
\end{equation}where $N$ denotes the batch size, $\hat{I}$ and $I$
denote the demoir\'{e}d image generated from the reconstruction
network and the ground truth image without moir\'{e}, respectively,
and $\epsilon$ is a parameter in charbonnier penalty and is set to
0.001 in our experiments.

\begin{table}
  \begin{center}
    \begin{tabular}{|l|c|c|c|c|}
      \hline
           & DnCNN & MSFE & Sun & Ours \\
      \hline
      \hline
      PSNR & 29.08     &    36.66  &  37.41   &  42.49    \\
      SSIM & 0.906     &    0.981  &  0.982   &  0.994    \\
      \hline
    \end{tabular}
  \end{center}
  \caption{Performance comparison with DnCNN~\cite{Zhang2016Beyond},MSFE~\cite{gao2019moire} and Sun~\cite{sun2018moire} on the LCDMoire~\cite{AIM19demoireDataset} validation set.}
  \label{table1}
\end{table}

\section{Experiment}
\label{section:experiment}
\subsection{Dataset and train detail}
\label{subsection:data_train}
We use the LCDMoire dataset~\cite{AIM19demoireDataset} provided by the AIM
contest as the training and validation sets, without using any
additional datasets. We use pytorch1.2 to build our proposed MDDM
model and use NVIDIA Titan V GPU with CUDA10.0 to accelerate
training. We use the Adam optimizer when training the model. We first
train a basic model which only contains three branches for feature
decomposition and demoir\'{e}ing. The initial learning rate is 1e-4,
and the learning rate is reduced by 10 times for every 30
epochs. Then we further increase the network size gradually to 4, 5
and 6 branches, and then finetune these models on the previous
basis. On the finetuning stage, the learning rate is set to 1e-5 and
is reduced by 10 times for every 50 epochs.

\subsection{Comparison with the state of the arts}
\label{subsection:compare}
In this section, we compare our proposed MDDM with some deep
learning-based demoir\'{e}ing methods~\cite{gao2019moire,sun2018moire}
proposed in recent years. Image demoir\'{e}ing is also related to
image denoising~\cite{Zhang2016Beyond}, and thus we compare our method
with DnCNN. For a fair comparison, all of the above methods are
retrained on the LCDMoire training set and are evaluated on the LCDMoire
validation set. We use the peak signal-to-noise ratio (PSNR) and
structural similarity (SSIM)~\cite{Zhou2004Image} as the performance
evaluation indicators. The result is shown in Table~\ref{table1}. From
Table~\ref{table1}, it can be seen that the results of our proposed
MDDM are significantly better than the state of the arts.

\subsection{Visual results}
\label{subsection:visual_res}
In this section, we compare the visual results of our proposed MDDM
with the above mentioned demoir\'{e}ing methods. The visual results
are shown in Figure~\ref{fig:compare_visualize}. The red square
means zoom-in of the image such that we can compare the details of
the results. From left to right are the moir\'{e} image, results of
DnCNN~\cite{Zhang2016Beyond}, MSFE~\cite{gao2019moire},
Sun~\cite{sun2018moire}, our proposed MDDM and the ground truth. PSNR
and SSIM are shown under the image. The visual results show that our
proposed MDDM is significantly better than other methods. Our MDDM
removes moir\'{e} patterns more cleanly, while other methods bring
more artifacts or can't completely remove moir\'{e} patterns.

\begin{table}[t]
\begin{center}
\begin{tabular}{|l|c|c|c|}
\hline
Scale         & PSNR & Parameter & Computation \\
\hline
\hline
1             & 27.71     &   \textless0.01    &  0.17      \\
1+2           & 27.89     &   0.77      &  187.17      \\
1+2+4         & 37.51     &   1.90      &  288.45      \\
1+2+4+8       & 38.22     &   3.54      &  358.68      \\
1+2+4+8+16    & 42.48     &   5.70      &  417.58      \\
1+2+4+8+16+32 & 42.49     &   8.01      &  472.38      \\
\hline
\end{tabular}
\end{center}
\caption{Performance (dB), parameters (M) and computation (GFLOPs) with different numbers of
  branches.}
\label{table2}
\end{table}

\begin{table}[t]
\begin{center}
\begin{tabular}{|l|c|c|c|}
\hline
Structure & PSNR & Parameter & Computation \\
\hline
\hline
No DFE    & 39.30     & 6.94         & 393.14     \\
DFE       & 42.49     & 8.01         & 472.38        \\
\hline
\end{tabular}
\end{center}
\caption{Performance (dB), parameters (M) and computation (GFLOPs) with and without dynamic
  feature encoding (DFE).}
\label{table:DFE}
\end{table}

\section{Ablation Study}
\label{section:ablation_study}
In this section, we study the structural design in our proposed
model. We analyze the modules we designed from the perspectives of
performance, model parameters, and computation. In Section~\ref{subsection:network_branch} we
conduct experiments on the number of branches. In Section~\ref{subsection:ablation_dfe}, we
analyze the performance gain brought by the DFE module, the
computational burden, and parameter growth.

\subsection{Network branches}
\label{subsection:network_branch}
Different branches learn image features at different resolutions and
encode image representations at different frequencies, and therefore
the number of branches significantly affects the performance of the
model. In this section, we study the relationship between model
performance and network branches. The results are shown in
Table~\ref{table2}. We use PSNR as an indicator of performance. We
also compare the number of parameters and the number of
single-precision floating point (FP32) calculations. The experimental
results are shown in Table~\ref{table2}.

\begin{figure}[t]
\begin{center}
\includegraphics[width=0.8\linewidth]{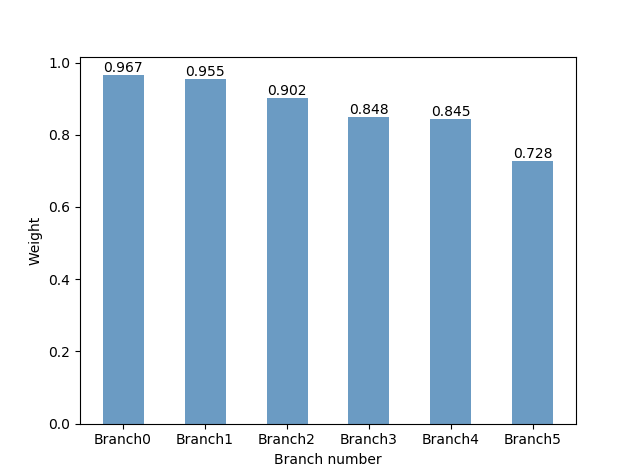}
\end{center}
\caption{Branch importance visualization.}
\label{fig:branch_importance}
\end{figure}

At the end of the network, we have designed a scaling module to
automatically learn the weight of each branch. These weights can be
viewed as the indicators of the importance of branches for the final
reconstruction or demoir\'{e}ing
image. Figure~\ref{fig:branch_importance} visualizes the weight of
each branch. We can see that branches with higher resolutions are of
greater importance.

\subsection{Dynamic feature encoding}
\label{subsection:ablation_dfe}
To cope with the dynamics of moir\'{e}, we use dynamic feature
encoding (DFE) branches to encode the dynamic properties of
moir\'{e}. We analyze the demoir\'{e}ing performance and model
complexity with and without DFE. The experimental results are shown in
Table~\ref{table:DFE}. Dynamic feature encoding (DFE) is a lightweight
module which will not bring much model parameters and computation. Table~\ref{table:DFE_complexity} shows the parameters and calculations growth after adding DFE to each branch.


\begin{table}
\begin{center}
\begin{tabular}{|l|c|c|}
\hline
Branch  & Parameter & Computation \\
\hline
\hline
Branch1 & 0.11          & 12.68      \\
Branch2 & 0.27          & 15.90      \\
Branch3 & 0.58          & 16.70       \\
Branch4 & 0.77          & 16.95       \\
Branch5 & 1.07          & 17.01      \\
\hline
\end{tabular}
\end{center}
\caption{Model parameters (M) and computation (GFLOPs) of each branch with DFE.}
\label{table:DFE_complexity}
\end{table}

\section{Conclusion}
\label{section:conclusion}
Digital cameras and mobile phones make photography more convenient in
life, but digital sensors often produce moir\'{e} when shooting scenes
with repetitive textures such as screens, which seriously affects the
quality of shooting. Moir\'{e} patterns are more variable and dynamic,
and therefore demoir\'{e}ing is more challenging than other image
restoration tasks. In response to this task, this work proposes a new
demoir\'{e}ing method, which includes two key elements: multi-scale
residual network and dynamic feature coding. Multi-scale networks
remove moir\'{e} in different bands and preserve more image
detail. Dynamic feature coding allows the model to adapt to the
changing nature of the moir\'{e}. It can be seen from the test results
that the model gain larger than 3dB in PSNR after adding the DFE
branch, and DFE is a lightweight module, which only brings a limited
increase in the number of parameters and computation. Through the
benchmark data, our proposed model can effectively remove moir\'{e}
and outperform the state of the arts.

\section{Acknowledgments}
\label{section:acknowledgements}
The authors would like to thank the editor and the anonymous reviewers 
for their critical and constructive comments and suggestions. This work 
was supported by the National Science Fund of China under Grant 
No. U1713208, Program for Changjiang Scholars.

{\small
\bibliographystyle{ieee_fullname}
\bibliography{egbib}
}

\end{document}